%% file: emnlp2022.tex
\pdfoutput=1

\documentclass[11pt]{article}

\usepackage[]{EMNLP2022}

\usepackage{times}
\usepackage{latexsym}

\usepackage[T1]{fontenc}

\usepackage[utf8]{inputenc}

\usepackage{microtype}

\usepackage{inconsolata}
\usepackage{microtype}
\usepackage{color}
\usepackage{latexsym}
\usepackage{amssymb}
\usepackage{amsmath}
\usepackage{multirow}
\usepackage{xspace}
\usepackage{enumitem}
\usepackage{graphicx}
\usepackage{epstopdf}
\usepackage{xspace}
\usepackage{array}
\usepackage{ragged2e}
\newcolumntype{P}[1]{>{\RaggedRight\hspace{0pt}}p{#1}}
\newcolumntype{X}[1]{>{\RaggedRight\hspace*{0pt}}p{#1}}
\usepackage{tcolorbox}
\usepackage{amsmath}
\usepackage{amsthm}
\usepackage{amssymb}
\usepackage{mathtools, nccmath}
\usepackage{wrapfig}
\usepackage{comment}
\usepackage{pifont}

\newcommand{\our}{\mbox{\textsc{STREAM}}\xspace}
\newcommand{\ourinit}{\mbox{$\textsc{STREAM}_{init}$}\xspace}
\newcommand{\ourhard}{\mbox{$\textsc{STREAM}_{hard}$}\xspace}

\newcommand{\task}{\mbox{\textit{named entity tagging}}\xspace}

\newcommand{\ie}{\textit{i.e.}}

\newcommand{\RA}{\textsc{TokenString}\xspace}
\newcommand{\RB}{\textsc{PreNGram}\xspace}
\newcommand{\RC}{\textsc{PostNGram}\xspace}
\newcommand{\RD}{\textsc{POSTag}\xspace}
\newcommand{\RE}{\textsc{DependencyRel}\xspace}
\newcommand{\EG}{\textit{``Thirty PD patients participated in the study''}\xspace}

\title{Distilling Task-specific Logical Rules from Large Pre-trained Models}

\author {
    Tao Chen\textsuperscript{\rm 1,3}\thanks{\quad Work done during an internship at Alibaba} ,
    Luxin Liu\textsuperscript{\rm 2},
    Xuepeng Jia\textsuperscript{\rm 2},
    Baoliang Cui\textsuperscript{\rm 2}, \\
    \textbf{Haihong Tang}\textsuperscript{\rm 2} \textbf{\&}
    \textbf{Siliang Tang}\textsuperscript{\rm 1,3}\thanks{\quad Corresponding author}
    \\
    \textsuperscript{\rm 1}Zhejiang University 
    \textsuperscript{\rm 2}Alibaba Group \\
    \textsuperscript{\rm 3}Alibaba-Zhejiang University Joint Research Institute of Frontier Technologies \\
    \texttt{\{ttc, siliang\}@zju.edu.cn} \\ \texttt{\{xique.llx, jiaxuepeng.jxp\}@alibaba-inc.com} \\
    \texttt{\{moqing.cbl, piaoxue\}@taobao.com}
}

\begin{document}
\maketitle

\input{sections/abstract}

\input{sections/introduction}

\input{sections/method}

\input{sections/exp}

\input{sections/related}

\input{sections/conclusion}

\input{sections/limitation}

\bibliography{anthology,custom}
\bibliographystyle{acl_natbib}

\end{document}

%% file: sections/abstract.tex
\begin{abstract}

Logical rules, both transferable and explainable, are widely used as weakly supervised signals for many downstream tasks such as named entity tagging.
To reduce the human effort of writing rules, previous researchers adopt an iterative approach to automatically learn logical rules from several seed rules.
However, obtaining more seed rules can only be accomplished by extra human annotation with heavy costs.
Limited by the size and quality of the seed rules, the model performance of previous systems is bounded.
In this paper, we develop a novel framework \our to distill task-specific logical rules from large pre-trained models.
Specifically, we borrow recent prompt-based language models as the knowledge expert to yield initial seed rules, and based on the formed high-quality instance pool that acts as an intermediary role, we keep teaching the expert to fit our task and learning task-specific logical rules.
Experiments on three public named entity tagging benchmarks demonstrate the effectiveness of our proposed framework.
With several predefined prompt templates, our system has gained significant improvements over previous state-of-the-art methods.
\end{abstract}

%% file: sections/introduction.tex
\section{Introduction}

\begin{figure}[t]
\centerline{\includegraphics[width=\linewidth]{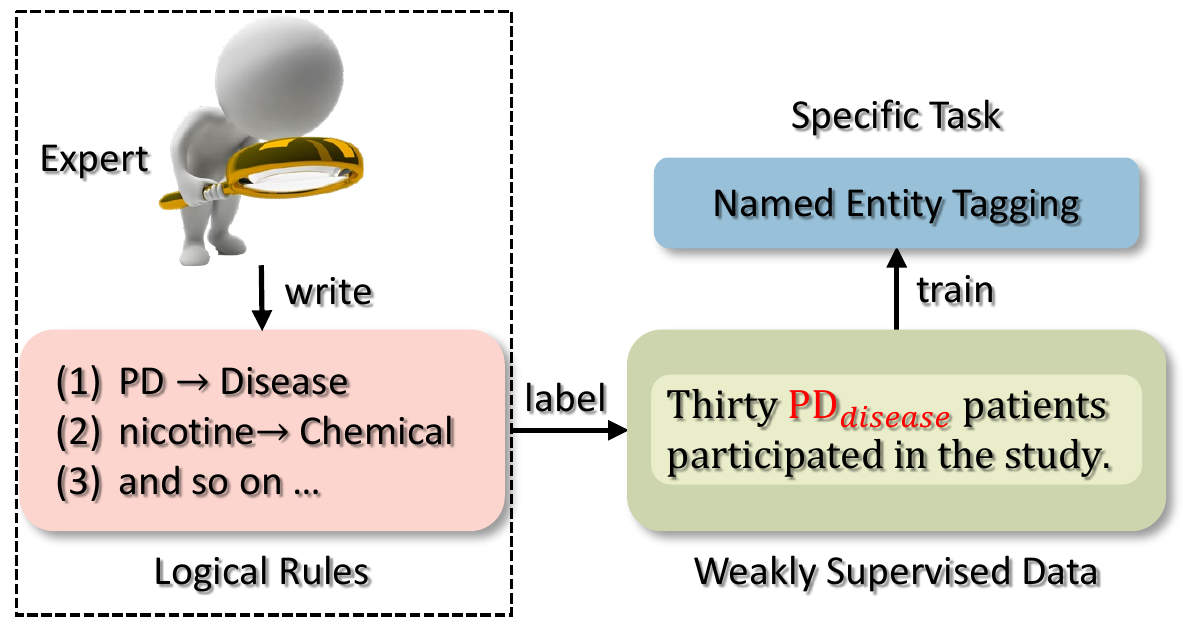}}
\caption{Schematic diagram of a typical rule-based weakly supervised named entity tagging system. Logical rules are used to label the data, and models can be trained on the weakly supervised data. Our goal in this work is to learn logical rules without any manual seed rules, corresponding to the dotted area in the figure.}
\label{task}
\end{figure}

Following the supervised learning paradigm, researchers resort to human annotation to obtain training data for specific tasks such as named entity tagging.
Though accurate, manual annotation construction is quite expensive and time-consuming.
In real scenarios, logical rules often serve as a source of weak supervision that provides abundant weakly supervised data for various downstream models, and compared with labeling data, applying rules can cover more application domains with better interpretability.
Therefore, rule-based weakly supervised systems~(Figure~\ref{task}) have attracted considerable attention in recent years.

In fact, it's not easy to develop an accurate and complete rule system, as the logical rules are usually summarized by human experts and the building process requires extensive domain knowledge.
Besides, there is no evaluation metric to guide annotators to select valuable rules. The usability and quality of acquired rules can not be guaranteed.
In this sense, how to build a reliable rule system with limited human effort is still an important challenge.

To solve above issue, previous researchers pay attention to the automatic construction of logical rules, which tends to start from a few seed rules and learn new logical rules by pre-defined similarity measures in an iterative manner.
Though proven to be effective, these systems still require manually constructed seed rules as the cold start.
Limited by human effort, the size of seed rules is usually small so that the system performance is bounded.

In this work, we propose an automated framework \our to di\underline{S}till \underline{T}ask-specific logical \underline{R}ules from large pr\underline{E}-tr\underline{A}ined \underline{M}odels.
Specifically,
(1) In order to get rid of the restrictions of the seed rules, we firstly ask large pre-trained models for help.
As the prompt-based pre-trained models own the zero-shot ability to generate candidate entity types, we design two appropriate prompt templates and achieve automatic acquisition of seed rules by the prompt model outputs' consistency.
(2) Once seed rules are obtained, we form a high-quality instance pool to train the downstream task, continuously add potential instances to the pool, and distill new logical rules from the pool in an iterative manner.
(3) Based on the convergent instance pool, we further fine-tune a new prompt-based model with more suitable prompt templates to obtain more reliable seed rules, and yield a better downstream task model.
Compared with previous methods, our system no longer relies on manual seed rules or dictionaries, but only needs several prompt templates.

Experiments on three public named entity tagging benchmarks demonstrate the effectiveness of our proposed framework \our, with consistent improvements over several baseline models and far exceed the state-of-the-art (SOTA) systems.
Besides, we perform a detailed ablation study to analyze the quality of our obtained seed rules, the convergence of our propose iterative framework, and some specific cases of learned logical rules.

Accordingly, the major contributions of our work are summarized as follows:

(1) We introduce the large pre-trained prompt-based models to end the dilemma that the logical rule learning systems require seed rules as a start.

(2) We develop an effective and stable framework to distill logical rules in an iterative manner, which combines prompt-based fine-tuning and rule distillation to achieve mutual enhancement.

(3) We conduct detailed experiments to illustrate the effectiveness and rationality of our framework --- with several predefined prompt templates, the performance of our method has surpassed previous rule learning systems based on manual rules.

%% file: sections/method.tex
\section{Methodology}

\subsection{Overview}

\begin{figure*}
\centerline{\includegraphics[width=\linewidth]{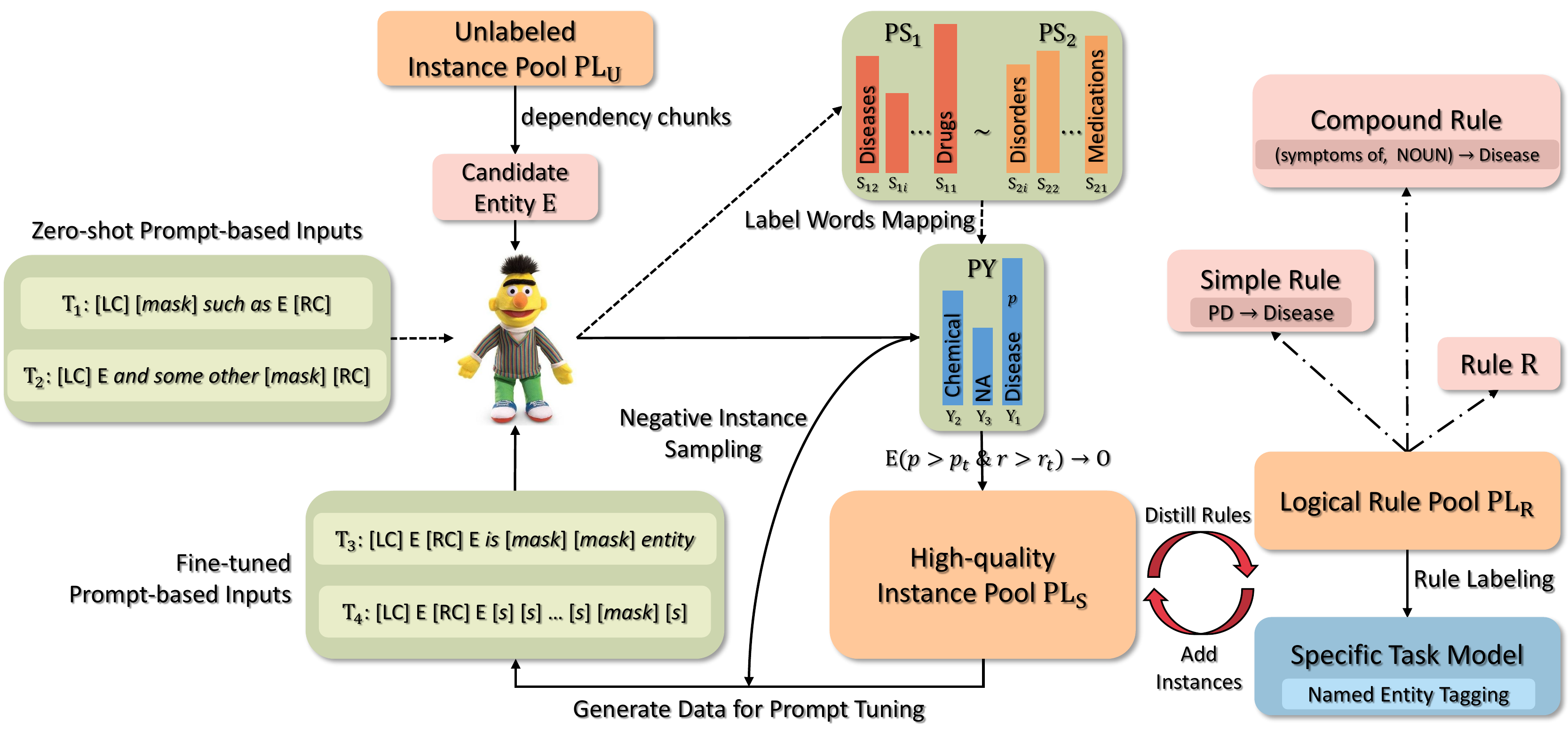}}
\caption{(1) In the first loop, zero-shot prompt models provide initial seed rules (\ie dotted line) and form a high-quality instance pool $\text{PL}_S$. (2) In the following loops, \our uses the positive instances from $\text{PL}_S$ and negative instances from sampling to fine-tune prompt models, generate better seed rules, and form a new instance pool $\text{PL}_S$. (3) During the process, the downstream model is trained and logical rules are distilled from the pool $\text{PL}_R$.}
\label{Model Framework}
\end{figure*}

In this work, we adopt named entity tagging as the specific downstream task to compare with previous work~\cite{li2021weakly} of learning logical rules.
The diagram of \our is visualized in Figure~\ref{Model Framework}.


\subsection{Logical Rules}

In real scenarios, logic rules can appear in various forms.
For convenience, we define the logical rules in the unified form of \textit{``if p then q}~(\ie \textit{p} $\rightarrow$ \textit{q})''.
In named entity tagging task, \textit{``p''} can be any logical expression and \textit{``q''} is the corresponding entity category.
For example, a logical rule may look like: ``\textit{if the entity's lexical string is PD\footnote{PD: Parkinson's disease}, then its corresponding entity label should be \textbf{disease}}''.

As demonstrated in previous work~\cite{zhou2002named}, we define five meta logical rules to tag named entities based on their lexical, contextual, and syntax information. In addition, some combinations of simple logical rules are also considered.

\subsubsection{Meta Logical Rules}

Following existing literature, our pre-defined meta-rules are:
(1) \RA rule matches entity's lexical string;
(2) \RB rule matches entity's preceding context tokens;
(3) \RC rule matches entity's succeeding context tokens;
(4) \RD rule matches entity's part-of-speech tags;
(5) \RE rule matches the dependency relations of the entity and its headword.

\begin{figure}[htbp]
\centerline{\includegraphics[width=\linewidth]{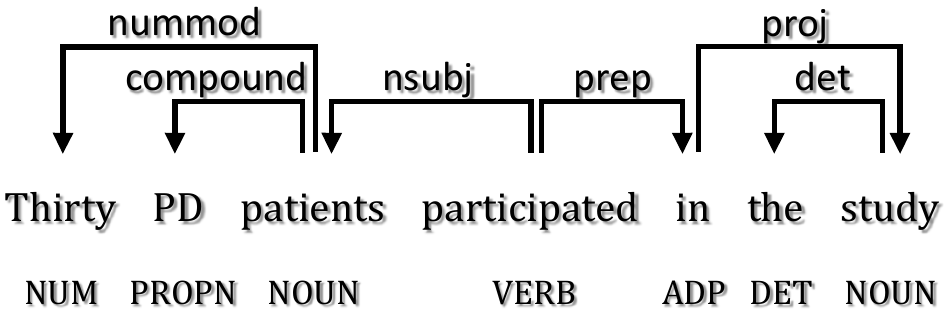}}
\caption{Dependency parsing example.}
\label{dependency diagram}
\end{figure}

Figure~\ref{dependency diagram} shows an example with its dependency structure.
In this sentence, word \textit{PD} is a potential disease entity and following logical rules may exist:

\vspace{-0.1cm}
\noindent
\begin{quote}
    \RA==\ \textit{PD} $\rightarrow$ disease \\
    \RB==\ \textit{thirty} $\rightarrow$ disease \\
    \RC==\ \textit{patients} $\rightarrow$ disease \\
    \RD==\ \small\textsc{PROPN}\normalsize\ $\rightarrow$ disease \\
    \RE==\ \\ (compound, \textit{patient}) $\rightarrow$ disease
\end{quote}
\vspace{-0.1cm}

In fact, above simple rules may sometimes fail to work, therefore we introduce complex rules, which combine several simple rules into compound rules by logical connectives including and ($\land$), or ($\lor$) and negation ($\lnot$).
For example, only a mention that satisfies both rule \RC==\ \textit{patients} and rule \RD==\ \small\textsc{PROPN}\normalsize\ can be a disease entity.

\subsubsection{Logical Rules Mining}
\label{rule mining}

After defining the form of meta logical rules, we traverse the entire training set and recall all potential rules that satisfy the format of meta rules.

\subsection{Zero-shot Prompt Models as Seed Rules}

In our proposed framework \our, we design a zero-shot prompt-based module to generate seed rules from pre-trained models without any manual seed rules, and the details are as follows.

\subsubsection{Zero-shot Prompt Model Inputs}
\label{sec: zero-shot prompt inputs}

For each sentence in the unlabeled training corpus $\text{PL}_{U}$, we first obtain its noun chunks $\text{E}$ by open-source dependency parsing tools such as \textit{Spacy}\footnote{https://spacy.io/models/en\#en\_core\_web\_sm}.
Actually, each noun chunk $\text{E}_i$ may be a potential entity, so that we construct the following prompt-based pattern proposed in~\cite{dai2021ultra} to mine its possible entity types:
\begin{equation}
    \text{T}_1:\text{LC} \dots \text{[}\textit{mask}\text{]}\ \textit{such}\ \textit{as}\ \text{E} \dots \text{RC}
\end{equation}
where $\text{LC}$ and $\text{RC}$ are left and right context tokens.
With the help of the zero-shot capability of large pre-trained models\footnote{In this work, we use RoBERTa-base as the prompt model.}, we can obtain the condition probability output ({$\text{PS}\subset \mathbb{R}^{V}$}, where $V$ is the vocabulary size) of noun chunk $\text{E}$ at the position of special token \text{[}\textit{mask}\text{]}.
For example, the prompt input of the above sentence \EG with noun chunk \textit{PD} is:
\begin{equation}
    \textit{Thirty}\ \text{[}\textit{mask}\text{]}\ \textit{such}\ \textit{as}\ \textit{PD}\ \textit{patients}\dots \textit{study}
\end{equation}
At the position of word [\textit{mask}], pre-trained models are able to directly output entity types like \textit{``diseases''}, \textit{``disorders''}, \textit{``conditions''} and so on, and we select \textit{Top-K} types ($\text{S}_{1}:\{\text{S}_{11},\text{S}_{12},\dots,\text{S}_{1K}\}$) from the condition probability output $\text{PS}_1$ as the candidate types of noun chunk $\text{E}_i$, where $\text{S}_{1i}$ is the entity type that outputs the $i_{th}$ highest confidence in the conditional probability $\text{PS}_1$.

However, the above prompt-based input breaks left context and focus more on right context, and we propose another prompt-based input to take left context also into consideration. It is defined as:
\begin{equation}
    \text{T}_2:\text{LC}\dots \text{E}\ \textit{and}\ \textit{some}\ \textit{other}\ \text{[}\textit{mask}\text{]} \dots \text{RC}
\end{equation}

For this prompt-based input, we can also obtain its \textit{Top-K} entity types ($\text{S}_{2}:\{\text{S}_{21},\text{S}_{22},\dots,\text{S}_{2K}\}$), and the final type set $\text{S}$ of noun chunk $\text{E}$ should be generated from two above type sets $\text{S}_1$ and $\text{S}_2$.

\subsubsection{Label Words Mapping}

To bridge the gap between model's output types $\text{S}$ and our task-specific categories $\text{Y}$, we design a module to find label words mapping automatically.

Specially, to find the label word mapping for target category $\text{Y}_i$, we count the co-occurrence between prompt model's output types $\text{S}_i$ and target type $\text{Y}_i$ in all $\text{S}$, and filter out the types with high support\footnote{The number of occurrences, called the support.} ($\text{S}_{ih}:\text{S}$).
Actually, the co-occurring types in $\text{S}_{ih}$ are often synonyms or aliases of the target type  $\text{Y}_i$.
For instance, we can find target category \textit{diseases} and type \textit{disorders} tend to appear together in \text{S}, which means there is a label word mapping: $\textit{disorders} \rightarrow \textit{disease}$.
We define the founded label mapping as $\mathcal{M}$, which maps all entity types in $\text{S}_{ih}$ to target type $\text{Y}_i$, this is defined as:
\begin{equation}
    \mathcal{M}: \text{S}_{ih} \rightarrow \text{Y}_i,\ \text{Y}_i \in \text{Y}
\end{equation}

\subsubsection{Zero-shot Seed Rules}
\label{sec: initial seed rules}

By the label word mapping $\mathcal{M}$, we can convert the initial condition probability output $\text{PS}_{1,2}$ to our task-specific condition probability $\text{PY}_{1,2} \subset \mathbb{R}^{v}$, where $v$ is the target category number.

$\text{PY}_1$ and $\text{PY}_2$ are the entity type predictions for noun chunk $\text{E}$ that come from two different prompt-based models.
Therefore, if the two models' predictions are similar and have no differences, the final entity type $\text{O}$ can be considered as the common output type of the two models. It is defined as:
\begin{equation}
    \text{O}=\left\{
\begin{aligned}
\text{Y}_{11}\quad& \textit{if}\quad \text{Y}_{11}=\text{Y}_{21},\\
\text{unk}\quad & \textit{otherwise.}
\end{aligned}
\right.
\end{equation}
where $\text{Y}_{1i}$ is the entity type with the $i_{\text{th}}$ largest model confidence $p_{1i}$ in $\text{PY}_1$, and \text{unk} means the entity type is unknown due to models' divergence.
For these chunks $\text{E}$ with determined entity type $\text{O}\neq \text{unk}$, we further filter out the chunks with high model confidence $p$, where $p=\min(p_{11},p_{21})$.

Besides, we also use a support threshold to further filter out high-quality chunks (\ie chunks that occur less often may be noisy), and we define the final obtained chunks pool as $\text{PL}_S$.

\begin{equation}
    \text{PL}_S=\{\text{E}:p>p_t, r>r_t\}
\end{equation}
where $p_t,r_t$ are confidence and support thresholds.
Actually, any noun chunk $\text{E}$ with entity category $\text{O}$ in the high-quality chunk pool $\text{PL}_S$ can be seen as an initial \RA rule: 
\begin{equation}
     \RA\ \text{==}\ \text{chunk E} \rightarrow \text{type O}
\end{equation}

\subsection{Distill Task-specific Logical Rules from High-quality Instance Pool}
\label{sec: learn rules}

Once seed rules are obtained, we can fetch all instances that matches the seed rules to form a instance pool $\text{PL}_S \rightarrow \text{PL}_{R}$.
Based on the initial pool $\text{PL}_{R}$, we aim to add high-quality instance to the pool, and distill new logical rules from the pool.

\subsubsection{Add High-quality Instances to the Pool}
\label{sec:add instances to the pool}

Based on the high-quality instances in $\text{PL}_{S}$, we can train specific~(\ie named entity tagging) downstream models, and the trained model is defined as $\mathcal{F}$.
After that, we use the trained model to generate a pseudo label for each unlabeled instance in $\text{PL}_{U}$.

To identify potential high-quality instances in unlabeled instances~$\text{PL}_{U}$ , we use the instances in high-quality pool $\text{PL}_{S}$ as a guide.
In detail, for any unlabeled sentence $s_u\in \text{PL}_{U}$, its pseudo label given by the trained model is $\text{Y}_u=\mathcal{F}(s_u)$.
We randomly sample a certain number of high-quality instances from $\text{PL}_{S}$ with the same entity label $\text{Y}_u$, and estimate the similarity between the instance $s_u$ and these sample instances. This is defined as:
\begin{equation}
    \text{S-score}(s_u)=\text{Medium}[\text{sim}(s_u,s_i)],\ s_i\in \text{PL}_{S}
\end{equation}
where \text{sim} is the function to measure the semantic similarity between sentence $s_u$ and the sampled sentence $s_i$, and \text{Medium} means that the final score is the median~(\ie avoid the influence of outliers) of all pair scores ($\text{sim}(s_u,s_i), s_i\in \text{PL}_S$).
Besides, to decide the score threshold of adding instances to the pool $\text{PL}_S$, we also randomly select instance $s_j$ from $\text{PL}_S$, and calculate the similarity score between the instance $s_j$ and the remaining instances $\text{PL}_{S}/s_j$. This process is defined as:

\begin{equation}
    \text{S-score}_{t}=\text{Medium}[\text{S-score}(s_j,\text{PL}_{S}/s_j)],\ s_j\subset \text{PL}_{S}
\end{equation}

\subsubsection{Distill Task-Specific Rules from the Pool}
\label{sec:distill rules from the pool}

With such a high-quality instance pool $\text{PL}_{S}$, our goal is to find all high-quality rules from potential rules set \text{R} mined in section~\ref{rule mining}.
For any rule $\text{R}_u \in \text{R}$, we define its confidence score as:
\begin{equation}
    \text{R-score}(\text{R}_u)=\frac{M_{\text{R}_u}}{N_{\text{R}_u}}\log_2 N_{\text{R}_u}
\end{equation}
where $N_{\text{R}_u}$ is the number of sentences that meets rule $\text{R}_u$ in the pool $\text{PL}_S$, and $M_{\text{R}_u}$ is the number of sentences that matches rule $\text{R}_u$ correctly (\ie the rule labelling result is consistent with the high-confidence label $\text{O}$).
Similarly, we also use the existing high-quality rules to determine the dynamic threshold for filtering out potential rules:
\begin{equation}
    \text{R-score}_{t}=\text{Medium}[\text{R-score}(\text{R}_i)],\ \text{R}_i \subset \text{PL}_R
\end{equation}

Accordingly, we keep repeating the steps defined in sections \ref{sec:add instances to the pool} and \ref{sec:distill rules from the pool} until pool $\text{PL}_S$ or $\text{PL}_R$ is no longer updated.
During this process, high-quality instances are gradually added to the instance pool $\text{PL}_S$, and corresponding high-quality logical rules are also produced.

\subsection{Fine-tuned Prompt Model as Seed Rules}
\label{sec:fine models}

In section~\ref{sec: zero-shot prompt inputs}, we propose to utilize zero-shot prompt-based models to generate initial seed rules, however, these rules are just a compromise at the time of data cold start (\ie w/o any weakly labeled data).
Once we have collected enough high-quality instances in pool $\text{PL}_S$, we can further adjust the prompt-based model to adapt to our specific task, and generate seed rules with higher quality.

\subsubsection{Fine-tuned Prompt Model Inputs}
\label{sec:fine tune prompt inputs}

To further fine-tune prompt models, we construct two new prompt-based inputs as follows:
\begin{equation}
\begin{aligned}
    \text{T}_3: &\text{LC}\ ...\ \text{E}\ ...\ \text{RC }\text{E}\ \textit{is}\ \text{[}\textit{mask}\text{]}\ \text{[}\textit{mask}\text{]}\ \textit{entity} \\
    \text{T}_4: &\text{LC}\ ...\ \text{E}\ ...\ \text{RC }\text{E}\ \text{[}\textit{s}\text{]}\ \text{[}\textit{s}\text{]}\dots \text{[}\textit{s}\text{]}\ \text{[}\textit{mask}\text{]}\ \text{[}\textit{s}\text{]}
\end{aligned}
\end{equation}
where $\text{[}\textit{s}\text{]}$ is the soft mask token. In the input template $\text{T}_3$, we aim to make the prompt-based model to output entity types at the positions of two [\textit{mask}] tokens, and its label words mapping is:
\begin{align}
\text{[}\textit{mask}\text{]}\quad\text{[}\textit{mask}\text{]} \quad &\nonumber \\
\textit{a/an}\quad \quad \ \ \text{O} \quad \quad &\rightarrow \quad \text{E}\textit{'s entity label is }\text{O}\nonumber \\
\textit{not}\ \ \ \quad \quad \textit{an} \ \ \ \quad &\rightarrow \quad \text{E}\textit{ is not an entity}
\end{align}
For example, if the output words at the positions of two mask tokens \text{[}\textit{mask}\text{]} are ``\textit{a disease}'', it means chunk E is an entity and its label is disease.

Besides, we also use a soft embedding prompt-based input scheme~(\ie $\text{T}_4$), which can implicitly learn word embeddings at the positions of \text{[}\textit{s}\text{]} through gradient propagation.
In this case, our goal is to constrain the model to output the target entity types O directly at the position of \text{[}\textit{mask}\text{]}.

Compared to the zero-shot prompt-based input proposed in section~\ref{sec: zero-shot prompt inputs}, the above two prompt-based inputs do not destroy the original sentence structure and promote the models to better understand the meaning of the entire sentence.

\subsubsection{Negative Instance Sampling}
\label{sec:negative samples}

However, instances in pool $\text{PL}_S$ are all positive sentences so that the models can not be trained only on pool $\text{PL}_S$.
To solve this issue, we sample some high-quality negative instances also based on the consistency of model outputs, this is defined as:
\begin{equation}
    \text{O}=\left\{
\begin{aligned}
\text{Y}_{11}\quad& \textit{if} \quad  \text{Y}_{11}=\text{Y}_{21},\\
\text{NA}\quad& \textit{otherwise if} \quad \text{S}_{11}=\text{S}_{21},\\
\text{unk}\quad & \textit{otherwise.}
\end{aligned}
\right.
\end{equation}
In short, negative samples (\ie \text{NA}) are the samples with same zero-shot prompt model outputs ($\text{S}_{11}=\text{S}_{21}$), but not with any target entity type ($\text{S}_{11}\not \subset \text{Y}$).

\subsubsection{Fine-tuned Seed Rules}

Based on the positive sentences provided by the pool $\text{PL}_S$ and negative instances sampled in section~\ref{sec:negative samples}, we can fine-tune prompt-based models with the inputs defined in section~\ref{sec:fine tune prompt inputs}.
Then, similar to the approach in section~\ref{sec: initial seed rules}, we use the fine-tuned models to predict pseudo labels for all unlabeled sentences, and select the sentences with high model confidence and high support as new seed rules.
Immediately afterward, our system will repeat the steps in sections~\ref{sec: learn rules} and~\ref{sec:fine models} to continuously distill more new rules, form a larger pool $\text{PL}_S$ and fine-tune a better prompt-based model.

%% file: sections/exp.tex
\section{Experiments}


\subsection{Benchmark}

\noindent\paragraph{BC5CDR}\cite{li2016biocreative} is constructed with BioCreative VCDR task corpus. It contains 500 train, 500 dev and 500 test PubMed articales, with 15,953 chemical and 13,318 disease entities.
\noindent\paragraph{CHEMDNER}\cite{krallinger2015chemdner} contains 10,000 PubMed abstracts with 84,355 chemical entities, in which the training/dev/test set contain 14,522/14,572/12,434 sentences respectively.

\noindent\paragraph{CONLL2003}\cite{sang2003introduction} consists of 14,041/3,250/3,453 sentences in the training/dev/test data split of Reuters news articles\footnote{Following previous work, type \text{MISC} is not considered.}.

\subsection{Model and Metric}

In our experiment, we use different weakly supervised approaches to label the manual set\footnote{Manual annotations can not be seen during training.} and obtain models on the set.
To compare the quality of the weakly supervised data generated by different methods, we evaluate the models with human labels and report corresponding model performance.
The evaluation metrics in our experiments include Precision(P), Recall(R) and F1-score(F$_1$).
In \our, we use the model proposed in~\cite{jiang2019generalizing} as the specific tagging model, and all our reported values are the average over five runs.

\begin{table*}
    \centering
    \begin{tabular}{l|c|ccc|ccc|ccc}
    \hline
    \multirow{2}{*}{Method} & \multirow{2}{*}{\begin{tabular}[c]{@{}c@{}}Need\\ Seed ?\end{tabular}} & \multicolumn{3}{c|}{BC5CDR}                                                      & \multicolumn{3}{c|}{CHEMDNER$^*$} &   \multicolumn{3}{c}{CONLL2003}                                         \\ \cline{3-11} 
                            &                                                                                        & \multicolumn{1}{c|}{P} & \multicolumn{1}{c|}{R} & \text{F}$_1$         & \multicolumn{1}{c|}{P} & \multicolumn{1}{c|}{R} & F$_{1}$ & \multicolumn{1}{c|}{P} & \multicolumn{1}{c|}{R} & F$_{1}$\\ \hline \hline
    Seed Rules              & \ding{51}                                                                                   & \multicolumn{1}{c|}{\textbf{94.09}}     & \multicolumn{1}{c|}{3.81}   & 7.33              & \multicolumn{1}{c|}{\textbf{91.60}}     & \multicolumn{1}{c|}{13.19}  & 23.07 & \multicolumn{1}{c|}{\textbf{95.77}}     & \multicolumn{1}{c|}{2.76}  & 5.36   \\
    Seed-Tagger       & \ding{51}                                                                                    & \multicolumn{1}{c|}{\underline{78.33}}     & \multicolumn{1}{c|}{21.60}  & 33.86             & \multicolumn{1}{c|}{84.18}     & \multicolumn{1}{c|}{21.91}  & 34.78 & \multicolumn{1}{c|}{72.57}     & \multicolumn{1}{c|}{24.68}  & 36.83    \\
    LinkedHMM               & \ding{51}                                                                                    & \multicolumn{1}{c|}{10.18}     & \multicolumn{1}{c|}{15.60}  & 12.32             & \multicolumn{1}{c|}{23.99}     & \multicolumn{1}{c|}{10.77}  & 14.86 & \multicolumn{1}{c|}{19.78}     & \multicolumn{1}{c|}{31.51}  & 24.30    \\
    HMM-Agg                 & \ding{51}                                                                                    & \multicolumn{1}{c|}{43.70}     & \multicolumn{1}{c|}{21.60}  & 29.00             & \multicolumn{1}{c|}{49.60}     & \multicolumn{1}{c|}{18.40}  & 26.80 & \multicolumn{1}{c|}{52.00}     & \multicolumn{1}{c|}{8.50}  & 14.60    \\
    CGExpan                 & \ding{51}                                                                                    & \multicolumn{1}{c|}{40.96}     & \multicolumn{1}{c|}{24.75}  & 30.86             & \multicolumn{1}{c|}{45.70}     & \multicolumn{1}{c|}{25.58}  & 32.80 & \multicolumn{1}{c|}{55.97}     & \multicolumn{1}{c|}{28.70}  & 37.95    \\
    AutoNER                 & \ding{51}                                                                                    & \multicolumn{1}{c|}{42.22}     & \multicolumn{1}{c|}{30.66}  & 35.52             & \multicolumn{1}{c|}{66.83}     & \multicolumn{1}{c|}{27.59}  & 39.05 & \multicolumn{1}{c|}{32.07}     & \multicolumn{1}{c|}{5.98}  & 10.08    \\
    Self-Training           & \ding{51}                                                                                    & \multicolumn{1}{c|}{73.69}     & \multicolumn{1}{c|}{29.55}  & 42.19             & \multicolumn{1}{c|}{\underline{85.06}}     & \multicolumn{1}{c|}{20.03}  & 32.42 & \multicolumn{1}{c|}{\underline{72.80}}     & \multicolumn{1}{c|}{24.83}  & 37.03    \\
    TALLOR                 & \ding{51}                                                                                    & \multicolumn{1}{c|}{66.53}     & \multicolumn{1}{c|}{\underline{66.94}}  & \underline{66.73} & \multicolumn{1}{c|}{48.34}     & \multicolumn{1}{c|}{\underline{52.56}}  & \underline{50.36} & \multicolumn{1}{c|}{64.29}     & \multicolumn{1}{c|}{\underline{64.14}}  & \underline{64.22}     \\ \hline
    \our                    & \ding{55}                                                                                     & \multicolumn{1}{c|}{72.47}     & \multicolumn{1}{c|}{\textbf{67.90}}  & \textbf{70.11}    & \multicolumn{1}{c|}{63.93}          & \multicolumn{1}{c|}{\textbf{55.13}}       &  \textbf{59.20} & \multicolumn{1}{c|}{69.92}          & \multicolumn{1}{c|}{\textbf{72.30}}       &  \textbf{71.09}        \\ \hline
    \end{tabular}
    \caption{
    \label{overall performance table}
    Model performances on BC5CDR, CHEMDNER, and CONLL2003. Bold and underline indicate the best and the second best scores, * means the reported result is our re-implementation of author-provide code.
    }
\end{table*}

\subsection{Baseline}

We select several recent weakly supervised methods to compare, including current SOTA systems.

\noindent\paragraph{Seed Rules} uses manually annotated seed rules to match set and evaluate the label performance.
\noindent\paragraph{Seed-Tagger} uses seed rules to label manual set and train the tagging models on the labeled data.
\noindent\paragraph{CGExpan}\cite{zhang2020empower} expands lexicons by language models. Following previous work, we use CGExpan to expand the size of human annotated \RA rules (\ie lexicons) to 1000.
\noindent\paragraph{AutoNER}\cite{shang2018learning} labels untyped terms automatically with a pre-defined dictionary. We use the best expanded lexicon from CGExpan as the dictionary. Both of the expanded lexicon and the mined phrases from AutoPhrase~\cite{shang2018automated} as untyped mined phrases.
\noindent\paragraph{LinkedHMM}\cite{safranchik2020weakly} proposes to utilize a generative model to aggregate noisy rules, and forms weak supervision signals to train the models. We use the best expanded lexicon from CGExpan as the tagging rules and the mined phrases from AutoPhrase as the linking rules.
\noindent\paragraph{HMM-Agg}\cite{lison2020named} introduces the hidden Markov models to generate weak labels by labeling functions. We use the best expanded lexicons from CGExpan as the labeling functions.
\noindent\paragraph{Self-Training} uses the seed rules to get initial teacher models and iterates the processes of generating pseudo labels for unlabeled data and training student models following self-training scheme.
\noindent\paragraph{TALLOR}\cite{li2021weakly} bootstraps high-quality logical rules to train a neural tagger in an iterative manner, with selected, the most frequent manually annotated seed rules as the input.

\subsection{Overall Performance}

We summarize the model performances of our \our and above mentioned baselines in Table~\ref{overall performance table}.
From the table, we can see:
(1) Method Seed Rules yield a high accuracy, however, this simple matching pattern lacks generalization ability and results in a low model recall.
(2) Similarly, the Self-training method starts from a small amount of seed data and has a good model accuracy, but its model recall is poor due to the limited data size.
(3) Lexicon expanded model CGExpan and AutoNER sacrifice a certain model accuracy in exchange for more balanced model performance.
(4) Previous SOTA system TALLOR can learn logical rules in an iterative manner and achieves competitive model F1-score. Since this system still relies on initial seed rules, its model performance is bounded.


\begin{figure}[htbp]
\centerline{\includegraphics[width=0.5\textwidth]{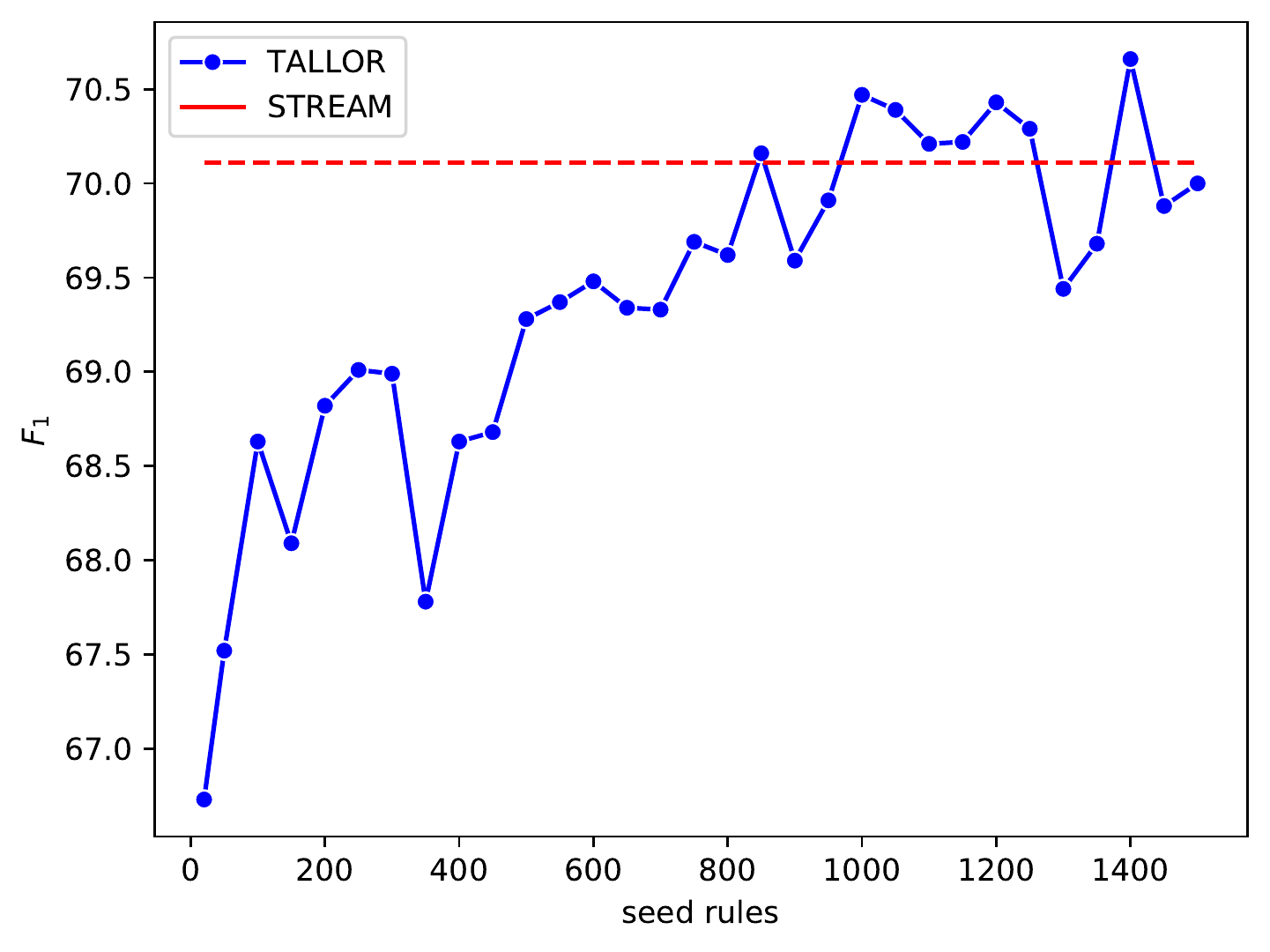}}
\vspace{-0.2cm}
\caption{Model Performances on dataset BC5CDR with different manually annotated seed rules size.}
\label{seed diagram}
\end{figure}

Compared to the above baseline models, our proposed system \our does not need any human-annotated seed rules or data.
We draw curves in Figure~\ref{seed diagram} to illustrate how many seed rules TALLOR needs to be comparable to our \our.
From the figure, we can see that:
When the number of most frequent seed rules reaches 825, the model performance of TALLOR exceeds \our for the first time.
However, acquiring such the most frequent seed rules is quite labor-intensive.

\subsection{Ablation Study}

To explore how our proposed system \our works, we now present ablation studies.

\noindent\paragraph{Initial Seed Rules}

In section~\ref{sec: initial seed rules}, we utilize zero-shot prompt-based models' consistent outputs as initial seed rules. We firstly conduct experiments to check the quality of initial seed rules. 

\begin{table}[htbp]
\centering
\scalebox{0.72}{
\begin{tabular}{l|ccc|ccc}
\hline
\multirow{2}{*}{Method} & \multicolumn{3}{c|}{BC5CDR}                                     & \multicolumn{3}{c}{CHEMDNER}                                    \\ \cline{2-7} 
                        & \multicolumn{1}{c|}{P}     & \multicolumn{1}{c|}{R}     & F$_1$    & \multicolumn{1}{c|}{P}     & \multicolumn{1}{c|}{R}     & F$_1$    \\ \hline \hline
TALLOR                  & \multicolumn{1}{c|}{66.53} & \multicolumn{1}{c|}{66.94} & 66.73 & \multicolumn{1}{c|}{48.34} & \multicolumn{1}{c|}{52.56} & 50.36 \\ \hline
\ourinit            & \multicolumn{1}{c|}{70.46} & \multicolumn{1}{c|}{64.99} & 67.62 & \multicolumn{1}{c|}{57.13} & \multicolumn{1}{c|}{50.52} & 53.62 \\
\our                  & \multicolumn{1}{c|}{72.47} & \multicolumn{1}{c|}{67.90} & 70.11 & \multicolumn{1}{c|}{63.93} & \multicolumn{1}{c|}{55.13} & 59.20 \\ \hline
\end{tabular}}
\caption{
\label{ablation1}
Ablation results of different rule learning systems on BC5CDR and CHEMDNER, method \ourinit uses the initial seed rules.
}
\end{table}
\vspace{-0.3cm}

We summarize the model performances of different logical rule learning systems in Table~\ref{ablation1}.
From the table we can see that:
With only the initial rules given by the zero-shot prompt model, our system \ourinit has surpassed the previous SOTA method TALLOR in metrics P and F$_1$, which means the generated seed rules are reliable.
Besides, we directly utilize the human-annotated labels to verify the quality of the initial rules in a more intuitive way:
When confidence threshold $p_t=0.3$ and support threshold $r_t=4$, about 219 seed rules are obtained with an accuracy of 98.6\%.

\begin{figure}[t]
\centerline{\includegraphics[width=0.5\textwidth]{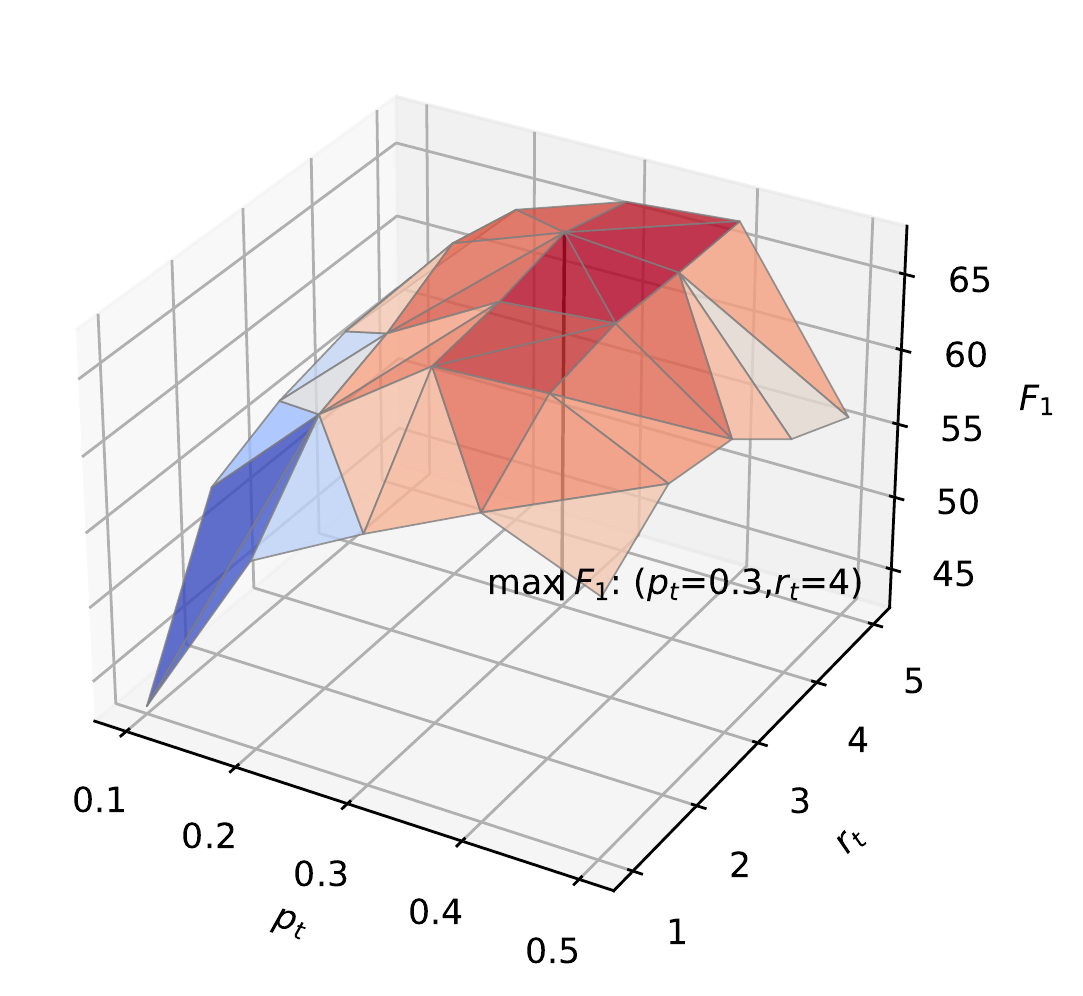}}
\vspace{-0.3cm}
\caption{Model performances of different hyperparameter combinations on dataset BC5CDR.}
\label{hyperparameter diagram}
\end{figure}

In the process of initial seed rules generation, hyperparameters $p_t$ and $r_t$ are relatively important. 
We draw the figure in Figure~\ref{hyperparameter diagram} to show the model performances of \ourinit with different combinations of $p_t$ and $r_t$.
From the figure, we can see that:
(1) When $p_t=0.3$ and $r_t=0.4$, our system \ourinit achieves the best performance.
(2) As hyperparameters $p_t$ or $r_t$ increases, the model performance first increases and then decreases. This is because a low parameter value may introduce some data noise, while a high parameter value may reduce the number of recalled rules.

\vspace{-0.3cm}

\noindent\paragraph{Fine-tuned Seed Rules}

After fine-tuning prompt-based models, the model can output high confidence scores for those correct samples.
Thus, to filter out them, we adopt $p_t$ as 0.99, and following previous experience, we adopt $r_t$ as 4.
With this combination, we can find that about 272 seed rules are recalled with an accuracy of 97.4\%.    

In section~\ref{sec:fine tune prompt inputs}, we propose two prompt-based inputs $\text{T}_3$ and $\text{T}_4$:
$\text{T}_3$ directly uses hard words to form the prompt sentence while $\text{T}_4$ introduces soft embeddings to learn during training process.
To compare the two prompt inputs, we present an ablation study, and the results are in Table~\ref{ablation2}.

\begin{table}[htbp]
\centering
\scalebox{0.72}{
\begin{tabular}{l|ccc|ccc}
\hline
\multirow{2}{*}{Method} & \multicolumn{3}{c|}{BC5CDR}                                     & \multicolumn{3}{c}{CHEMDNER}                                    \\ \cline{2-7} 
                        & \multicolumn{1}{c|}{P}     & \multicolumn{1}{c|}{R}     & F$_1$    & \multicolumn{1}{c|}{P}     & \multicolumn{1}{c|}{R}     & F$_1$    \\ \hline \hline
TALLOR                  & \multicolumn{1}{c|}{66.53} & \multicolumn{1}{c|}{66.94} & 66.73 & \multicolumn{1}{c|}{48.34} & \multicolumn{1}{c|}{52.56} & 50.36 \\ \hline
\ourhard            & \multicolumn{1}{c|}{73.13} & \multicolumn{1}{c|}{65.82} & 69.28 & \multicolumn{1}{c|}{61.49} & \multicolumn{1}{c|}{55.65} & 58.42 \\
\our                  & \multicolumn{1}{c|}{72.47} & \multicolumn{1}{c|}{67.90} & 70.11 & \multicolumn{1}{c|}{63.93} & \multicolumn{1}{c|}{55.13} & 59.20 \\ \hline
\end{tabular}}
\caption{
\label{ablation2}
Ablation results of different prompt-based inputs on BC5CDR and CHEMDNER, method \ourhard uses the prompt-based input $\text{T}_3$, method \our uses the prompt input $\text{T}_4$.
}
\end{table}

\begin{table*}[htbp]
\centering
\scalebox{0.92}{
\begin{tabular}{l|l|l} 
\hline
Logical Rule Matched Sentence  & Logical Rule Condition $p$    & Entity Type $q$                                                                                                    \\
\hline \hline
\begin{tabular}[c]{@{}l@{}}Grade less than or equal to 2 \textbf{nausea$_\text{{\textcolor{red}{NOUN}}}$} \textbf{\textcolor{red}{and}} \\ \textbf{\textcolor{red}{vomiting}} occurred in 66\%, courses and phlebitis ... \end{tabular}                & \begin{tabular}[c]{@{}l@{}}\RC == \textbf{\textcolor{red}{and vomiting}}\\$\land$~\RD == \small\textbf{\textcolor{red}{NOUN}}\end{tabular} & $p \rightarrow q: \textbf{disease}$               \\ 
\hline
\begin{tabular}[c]{@{}l@{}}This study describes neuropsychiatric side effects in \\ patients \textbf{\textcolor{red}{after treatment with}} \textbf{mefloquine$_\text{{\textcolor{red}{PROPN}}}$}.\end{tabular}                                       & \begin{tabular}[c]{@{}l@{}}\RB == \textbf{\textcolor{red}{after treatment}} \\ \textbf{\textcolor{red}{with}} $\land$~\RD == \small\textbf{\textcolor{red}{PROPN}}~ ~~\end{tabular} & $p \rightarrow q: \textbf{chemical}$  \\ \hline
\begin{tabular}[c]{@{}l@{}}Prophylactic use of lamivudine with chronic immun-\\ osuppressive \textbf{\textcolor{red}{therapy for}} \textbf{rheumatologic disorders}\textbf{\textcolor{red}{.}}  \end{tabular}                                       & \begin{tabular}[c]{@{}l@{}}\RB == \textbf{\textcolor{red}{therapy for}} \\ $\land$~\RC == \small\textbf{\textsc{\textcolor{red}{[END]}}}\end{tabular} & $p \rightarrow q: \textbf{disease}$  \\

\hline
\end{tabular}}
\vspace{-0.1cm}
\caption{Cast study of learned logical rules on dataset BC5CDR, [END] means the end (.) of sentences.}
\label{case study}
\end{table*}

From the table we can see that:
(1) Whether template $\text{T}_3$ or $\text{T}_4$ is used, our system \our can achieve SOTA model performance.
(2) Soft template $\text{T}_4$ performs better because it can learn a more efficient prompt pattern during training process.

\noindent\paragraph{High-quality Instance Pool}

\begin{figure}[htbp]
\centerline{\includegraphics[width=0.5\textwidth]{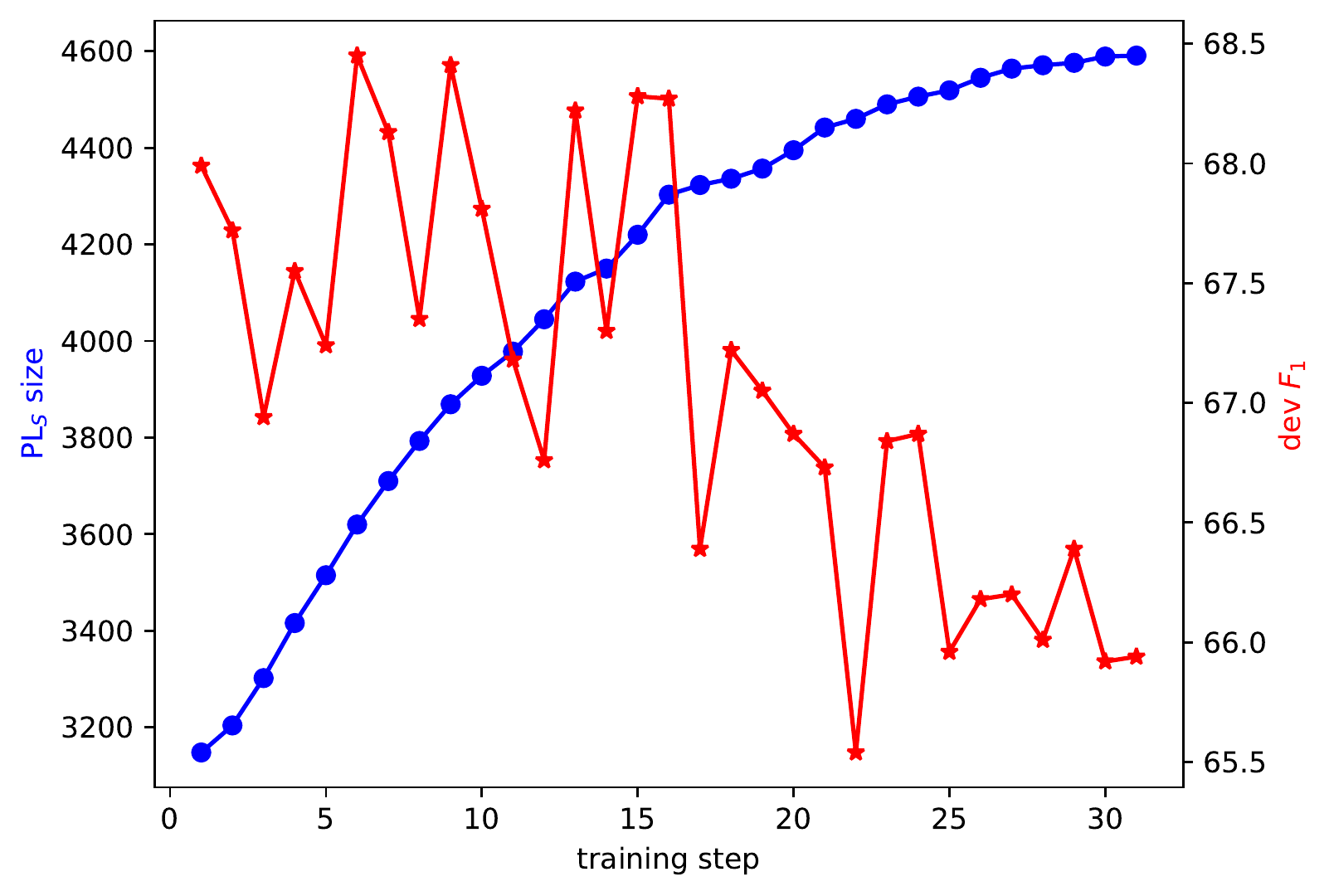}}
\vspace{-0.1cm}
\caption{Changes of $\text{PL}_S$ size and model performance on development set~(\ie dev $F_1$) during training process.}
\vspace{-0.3cm}
\label{change diagram}
\end{figure}

In section~\ref{sec: learn rules}, we add high-quality instances to $\text{PL}_S$ and distill new logical rules from $\text{PL}_S$ in an iterative manner.
To figure out how $\text{PL}_S$ changes during the training process, we now show the ablation in Figure~\ref{change diagram}.

From the figure, we can see:
(1) During the iterative process, the size of $\text{PL}_S$ continues to increase, but its growth rate gradually slows down.
(2) At the beginning of the iteration, high-quality instances are added to the pool $\text{PL}_S$, and the model performance increases.
However, at the later iteration stage, some noise instances may be introduced, which causes the model performance to decrease.

\subsection{Case Study: Learned Logical Rules }


In Table~\ref{case study}, we select three sentences and corresponding learned logical rules from the training corpus. For example:
(1) The entity mention in the first sentence matches the \RD form of \small\textbf{NOUN}\normalsize, and its \RC words are \textbf{and vomiting}, therefore, \our can infer a rule: (\RC == \textbf{and vomiting} $\land$ \RD == \small{\textbf{NOUN}}\normalsize) $\rightarrow$ \textbf{disease}. 
(2) In the third sentence, the \RB words of mention \textbf{rheumatologic disorders} are \textbf{therapy for} while the mention is just the end of sentence. \our can extract logical rule (\RB == \textbf{therapy} for $\land$ \RC == \small\textbf{END}\normalsize) $\rightarrow$ \textbf{disease} from this sentence.

%% file: sections/related.tex
\section{Related Work}

\noindent\paragraph{Weak Supervision} 

To alleviate the issue of limited labeled data, previous researchers made many efforts to improve \task systems from different perspectives:
(1)~\citep{ren2015clustype,fries2017swellshark,giannakopoulos2017unsupervised} introduce distant supervision~\cite{mintz2009distant}, an automated method to label data by aligning text with remote knowledge bases, to build NER systems without human supervision.
(2)~\citep{shang2018automated} uses typed lexicons and~\citep{peng2019distantly} uses incompetent dictionaries as the indirect supervision to guide model training.
However, lexicon or KB is not always available and its construction is expensive.
(3)~\citep{niu2003bootstrapping,huang2010inducing} use classical bootstrap methods to build NER systems.
(4)~\citep{lin2020triggerner} introduces ``entity triggers'', an effective proxy of human explanations for facilitating label-efficient learning of NER models.
(5) Recently,~\citep{bach2017learning,lison2020named,safranchik2020weakly} focus on rule aggregation to learn from noisy supervision, and~\citep{li2021weakly} proposes to learn logical rules from selected seed rules to generate more diverse pseudo labels, and achieves the SOTA model performance.
However, the above systems still rely on some manual data or rules, while our system yields rules from pre-trained models under limited prompt patterns.

\noindent\paragraph{Language Models}
~\citet{vaswani2017attention} proposed a self-attention based architecture --- Transformer, and it soon becomes the backbone of many following language models.
By pre-training on a large-scale corpus, BERT~\citep{devlin-etal-2019-bert} obtains the ability to capture a notable amount of ``common-sense'' knowledge and gains significant improvements on many tasks following the fine-tune scheme. 
Recently,~\citep{gao2020making,han2021ptr,wei2021finetuned} found that the prompt-based models achieve remarkable few-shot performance, and reformulate the traditional paradigm of fine-tuning to prompt-tuning, which could better utilize the knowledge of the pre-trained models. 

%% file: sections/conclusion.tex
\section{Conclusion}

In this work, we propose an automated framework \our to distill task-specific logical rules from large pre-trained models.
Experiments show the effectiveness of \our, with stable and significant improvements over different baseline models.
